\documentclass[letterpaper]{article} 
\usepackage{aaai2026}  
\usepackage{times}  
\usepackage{helvet}  
\usepackage{courier}  
\usepackage[hyphens]{url}  
\usepackage{graphicx} 
\usepackage{natbib}  
\usepackage{caption} 
\usepackage{booktabs}
\usepackage{algorithm}
\usepackage{algorithmic}
\usepackage{multirow}
\usepackage{gensymb}
\usepackage{enumitem}
\usepackage{parskip}
\usepackage{textcomp}
\usepackage{makecell}
\usepackage{newfloat}
\usepackage{listings}
\DeclareCaptionStyle{ruled}{labelfont=normalfont,labelsep=colon,strut=off} 
\floatstyle{ruled}
\newfloat{listing}{tb}{lst}{}
\floatname{listing}{Listing}
\title{\textbf{\textit{PANORAMA}}: \\ The Rise of Omnidirectional Vision in the Embodied AI Era}
\author{
    Xu Zheng\textsuperscript{\rm 1,\rm 2,\dag},
    Chenfei Liao\textsuperscript{\rm 1,\rm 3,*},
    Ziqiao Weng\textsuperscript{\rm 1,*},
    Kaiyu Lei\textsuperscript{\rm 1,*},
    Zihao Dongfang\textsuperscript{\rm 1},
    Haocong He\textsuperscript{\rm 3}, \\
    Yuanhuiyi Lyu\textsuperscript{\rm 1},
    Lutao Jiang\textsuperscript{\rm 1},
    Lu Qi\textsuperscript{\rm 6,7},
    Li Chen\textsuperscript{\rm 1},
    Danda Pani Paudel\textsuperscript{\rm 2}, 
    Kailun Yang\textsuperscript{\rm 5}, \\
    Linfeng Zhang\textsuperscript{\rm 3}, 
    Luc Van Gool\textsuperscript{\rm 2}, 
    Xuming Hu\textsuperscript{\rm 1,\rm 4,\ddag} 
}
\affiliations{
    \textsuperscript{\rm 1}HKUST(GZ), 
    \textsuperscript{\rm 2}INSAIT, Sofia University “St. Kliment Ohridski”, 
    \textsuperscript{\rm 3}Shanghai Jiao Tong University,\\
    \textsuperscript{\rm 4}CSE, HKUST,
    \textsuperscript{\rm 5}Hunan University,
    \textsuperscript{\rm 6}Wuhan University,
    \textsuperscript{\rm 7}Insta360 \\
    \textsuperscript{\rm \dag}Presenter,
    \textsuperscript{\rm *}Core Contributor,
    \textsuperscript{\rm \ddag}Corresponding Author \\

    \{zhengxu128, lcfgreat624\}@gmail.com
}

\usepackage{bibentry}

\begin{document}

\maketitle

\begin{abstract}

Omnidirectional vision—using 360$^{\circ}$ vision to understand the environment—has become increasingly critical across domains like robotics, industrial inspection, and environmental monitoring. 
Compared to traditional pinhole vision, omnidirectional vision provides holistic environmental awareness, significantly enhancing the completeness of scene perception and the reliability of decision-making.
However, foundational research in this area has historically lagged behind traditional pinhole vision. This talk presents an emerging trend in the embodied AI era: the rapid development of omnidirectional vision, driven by growing industrial demand and academic interest.
We highlight recent breakthroughs in omnidirectional generation, omnidirectional perception, omnidirectional understanding, and related datasets.
Drawing on insights from both academia and industry, we propose an ideal panoramic system architecture in the embodied AI era: PANORAMA, which consists of four key subsystems. 
Moreover, we offer in-depth opinions related to emerging trends and cross-community impacts at the intersection of panoramic vision and embodied AI, along with the future roadmap and open challenges.
This talk synthesizes state-of-the-art advancements and outlines challenges and opportunities for future research in building robust, general-purpose omnidirectional AI systems in the embodied AI era.
\end{abstract}


\section{Motivation \& Problem Space}
The pursuit of artificial visual perception that rivals human capabilities has long been a cornerstone of computer vision~\cite{zhao2024review,zheng2023deep,zheng2025retrieval}. 
Decades of research focused on pinhole-based visual perception, which offers a narrow, frustum-limited view of the world~\cite{zheng2024360sfuda++,lyu2024unibind}. 
In the past eras, pinhole-based visual perception has driven significant advancements in fields like image classification~\cite{chen2021review}, object detection~\cite{kaur2023comprehensive}, and semantic segmentation~\cite {mo2022review}.
While in the current era of embodied AI, more complex tasks, such as indoor/outdoor navigation~\cite{wang2025panogen++}, are emerging.
Such tasks rely on the comprehensive and thorough perception of environments,  requiring a holistic, 360$^{\circ}$ understanding of their surroundings~\cite{xu2024survey,xu2025embodied}. 
Therefore, omnidirectional vision has gradually become a more competitive solution than pinhole-based vision in such an embodied AI era~\cite{ai2025survey}.

The integration of panoramic visual technology with embodied intelligence faces fundamental gaps that define the core problem space.
From a systematic perspective, these gaps can be categorized into three types as shown in Figure~\ref{fig:figure1}: data bottlenecks, model capabilities, and application blanks:

\begin{figure}[t!]
    \centering
    \includegraphics[width=1\linewidth]{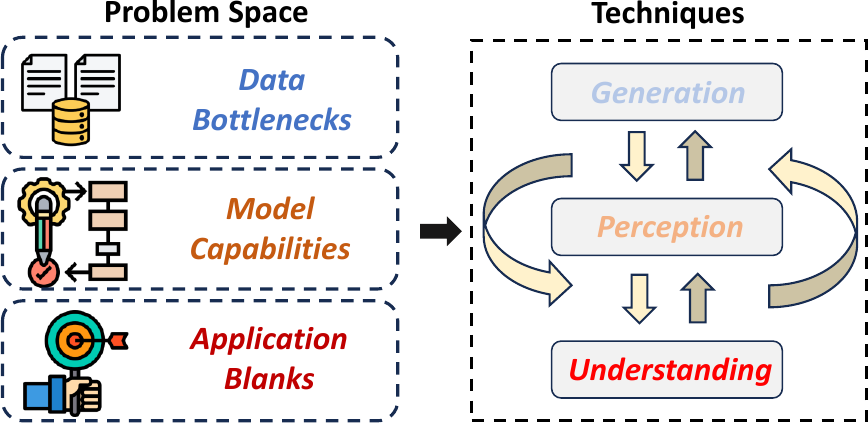}
    \caption{Challenges \& techniques of 360 vision.}
    \label{fig:figure1}
\end{figure} 
    
\noindent \textbf{Data Bottlenecks:}
Panoramic images, which are often distorted due to equirectangular projection (ERP) and typically captured at high resolutions, are more costly to annotate manually compared to other image types, such as pinhole images. The geometric distortions inherent in these projections make conventional automated annotation tools, typically designed for pinhole images, ineffective. These challenges significantly hinder the development of large-scale, high-quality datasets~\cite{li2024omnissrzeroshotomnidirectionalimage, zhang2025towards, Jiang_2025, huang2024360loc, park2024360,liu2025shifting}, creating a data bottleneck that impedes the advancement of omnidirectional vision in the field of embodied AI.

\noindent \textbf{Model Capabilities:}
From the modeling perspective, most existing pre-trained models encode inductive biases (e.g., translation invariance) through operations like convolution and pooling, which are applicable for pinhole images. 
However, these models are incapable of understanding the distortion characteristics of panoramic images, leading to a significant decline in performance when the models are directly transferred to deal with panoramic images. 
Thus, how to efficiently bridge the domain gap of data at the model level has become a key issue in the research on panoramic vision~\cite{coors2018spherenet,shen2022panoformer,su2019kernel,yun2023egformer}, especially in the era of embodied AI.

\noindent \textbf{Application Blanks:} 
When new sensors (360 cameras) meet a new era (embodied AI), many original scenarios, such as industrial scenarios, are likely to experience new technological iterations. 
However, different application scenarios exhibit distinct characteristics and priorities, which have spawned numerous scenario-specific research sub-fields.
This has given rise to numerous sub-fields of research that are scenario-specific.
Due to the lack of interdisciplinary talents and the insufficiency of existing panoramic data and models, these sub-fields, such as panoramic production safety inspection, panoramic forest fire detection, and so on, currently lack sufficient exploration~\cite{yan2024panovos,jung2025edm,fu2025segman}.

In summary, only by systematically tackling the threefold challenges of data, models, and applications can we truly unlock the enormous potential of omnidirectional vision in the era of embodied AI, thereby laying a solid foundation for achieving general and robust embodied intelligence.
    
\section{Recent Technical Advances} 
\begin{table*}[h!]
\renewcommand{\arraystretch}{1.5}
\centering
\scriptsize
\resizebox{\textwidth}{!}{%
\begin{tabular}{@{}c c c c c c@{}}
\toprule
\textbf{Scene} & \textbf{Source} & \textbf{Dataset} & \textbf{Year} & \textbf{Annotations} & \textbf{Scale} \\
\midrule
\multirow{13}{*}{Indoor} & \multirow{6}{*}{Real} & PanoContext~\cite{dong2024panocontext} & 2014 & Layouts, object 3D bounding boxes & 700 full-view panoramas \\
& & Stanford 2D-3D-S~\cite{armeni2017joint} & 2017 & RGB, depth, semantics &  70,000 images \\
& & Matterport3D~\cite{chang2017matterport3d} & 2017 & RGB-D panoramas, 2D/3D semantics \& poses & 10.8 K images, 90 buildings \\
& & ZInD~\cite{cruz2021zillow} & 2021 & 360$^{\circ}$ panoramas, 3D layouts, floorplans & 71 K panoramas, 1.5 K homes \\
& & HM3D~\cite{ramakrishnan2021habitat} & 2021 & Textured 3D mesh reconstructions of interiors & 1000 building-scale scenes \\
& & ToF-360~\cite{kanayama2025tof} & 2025 & 2D/3D Segmentation, Layout Estimation & 207 total panoramas \\
\cline{2-6}
& \multirow{7}{*}{Synthetic} & 3D60~\cite{Zioulis_2018_ECCV} & 2018 & Rendered panoramas + depth & ~60 real-world scenes \\
& & Structured3D~\cite{zheng2020structured3d} & 2020 & CAD-rendered panoramas + geometry/semantics & ~3.5 K layouts \\
& & DeepPanoContext~\cite{zhang2021deeppanocontext} & 2021 & Synthetic panoramas with layout, object shape, pose & Rich 3D annotations \\
& & SynPASS~\cite{10546335} & 2024 & 22-class semantic synthetic panoramas & ~9 K images \\
& & HM3DSem~\cite{Yadav_2023_CVPR} & 2022 & Dense object instance semantic annotations & 142,646 object instances \\
& & VLN-RAM~\cite{wei2025unseen} & 2024 & Human-aware visual language navigation
& 90 scenes, 145 human activity descriptions \\
& & OSR-Bench~\cite{dongfang2025multimodal} & 2025 & QA of omnidirectional spatial reasoning & 153,000+ diverse question-answer pairs \\
\midrule
\multirow{9}{*}{Outdoor} & \multirow{6}{*}{Real} & StreetLearn~\cite{mirowski2019streetlearn} & 2019 & StreetView panoramas + connectivity graph & ~143 K images \\
& & CVIP360~\cite{mazzola2021dataset} & 2021 & 360$^{\circ}$ video + pedestrian bounding boxes & 16 videos \\
& & 360VOT~\cite{huang2023360vot} & 2023 & 360$^{\circ}$ video, tracking + segmentation labels & 120 sequences, 113 K frames \\
& & 360 in the Wild~\cite{park2024360} & 2024 & 360$^{\circ}$ images + poses + depth & 25 K real scenes \\
& & 360Loc~\cite{huang2024360loc} & 2024 & 360$^{\circ}$ images + 6DoF poses, cross-device & Multi-device localization testbed \\
& & Dense360~\cite{zhou2025dense360} & 2025 & Entity-grounded panoramic scene descriptions & 160K panoramas + 5M entity-level captions \\
\cline{2-6}
& \multirow{2}{*}{Synthetic} & OmniScape~\cite{sekkat2020omniscape} & 2020 & Multiple projection panoramas + semantics & 10,000 captures \\
& & OmniHorizon  \cite{Bhanushali_2024_CVPR}& 2023 & Synthetic outdoor panoramas + depth/normal & 24,335 images  \\
\midrule
\multirow{2}{*}{UAV (Flight)} & \multirow{1}{*}{Real} & UAV-ERP~\cite{zhang2022classification} & 2022 & ERP panoramas captured & ~2.3 K images annotated \\
\cline{2-6}
& \multirow{1}{*}{Synthetic} & AirSim-360~\cite{airsim2017fsr} & 2018 & Unreal-simulated panoramas + depth & Configurable generation via API \\
\bottomrule
\end{tabular}}
\caption{Overview of representative omnidirectional datasets.}
\label{tab:omni-dataset}
\end{table*}

Recent advances in omnidirectional vision can be categorized into three closely related areas: generation, perception, and understanding. In the area of generation, existing research focuses on creating structurally consistent panoramic images and addressing domain-specific challenges, such as those posed by spherical projections. In perception, the focus is on adapting models designed for pinhole image tasks to handle omnidirectional perception tasks. In the area of understanding, the emphasis is on enabling models, particularly multimodal large language models, to interpret panoramic images, with a particular focus on extracting and understanding the spatial information they contain.
    
\subsection{Omnidirectional Generation}
Researchers focus on the generative adversarial network-based methods for omnidirectional generation at the early stage~\cite{ai2024dream360,chen2022text2light,cheng2022inout,wang2022stylelight,oh2022bips}. 
Typically, through a two-stage strategy, including both codebook-based panorama outpainting and frequency-aware refinement, Dream360~\cite{ai2024dream360} successfully generates high-quality and high-resolution panorama images based on selected viewports. 
As diffusion models have become the mainstream method in the field of generation recently, related research has gradually emerged~\cite{li2023panogen,li20244k4dgen,wang2024360dvd,wang2023360,wang2024customizing,ye2024diffpano,li2024omnidrag,wu2023panodiffusion}.
Among these works, PanoDiffusion~\cite{wu2023panodiffusion} generates panorama images through a two-branch diffusion structure, which allows RGB-D data as inputs during training. Thus, more spatial information can be injected into the generation model, improving the quality of the generated panoramic images. Meanwhile, OmniDrag~\cite{li2024omnidrag} controls the generation of panoramic images based on trajectories, further enhancing the user-friendliness of panoramic image generation.
    
\subsection{Omnidirectional Perception} 
Considering the data bottleneck problems of omnidirectional vision, the domain adaptation technique has become a popular solution that enables models to deal with panoramic images with unlabeled data~\cite{zheng2024360sfuda++}.
The existing strategies can be primarily classified into three types: adversarial learning-based strategy, pseudo-label-based strategy, and prototype-based strategy~\cite{zhong2025omnisam}. 
Adversarial learning-based strategies introduce a discriminator to force the model to generate features that are difficult to distinguish between domains, thereby capturing domain-invariant representations. 
Pseudo-label-based strategies train models by generating self-supervised labels for the target domain data, where GoodSAM~\cite{zhang2024goodsam} and GoodSAM++~\cite{zhang2024goodsam} refine pseudo-labels using the Segment Anything Model (SAM)~\cite{kirillov2023segment} to provide more reliable pseudo-labels for the target model. 
Meanwhile, OmniSAM~\cite{zhong2025omnisam} proposes a dynamic pseudo-label updating mechanism to improve the credibility of the pseudo-label.
Prototype alignment strategies aim to align the centers of high-level features between the source and target domains to reduce domain discrepancies. Previous works, such as 360SFUDA++~\cite{zheng2024360sfuda++} and OmniSAM~\cite{zhong2025omnisam}, center on matching the distortion and abstracting the semantics with prototypes, yielding significant improvements.

\subsection{Omnidirectional Understanding} 
Current multi-modal large language models tend to be trained through normal images, especially the pinhole images. 
Consequently, these models struggle to understand panoramic images, having never encountered them before.
From the perspective of data, recent works focus on building omnidirectional understanding datasets and benchmarks~\cite{dongfang2025multimodal,zhang2025towards,song2024video,zhou2025dense360,chou2020visual}. 
Especially, OSR-bench~\cite{dongfang2025multimodal} creates a concept of cognitive maps, dividing the whole panoramic image into patches and labeling patches by the objects within them.
Through this hierarchical approach, OSR-Bench achieves fast and effective data labeling and benchmarking.
Meanwhile, OmniVQA~\cite{zhang2025towards} achieves efficient data labeling through agent collaboration. 
While from the perspective of models, current methods tend to apply the GRPO techniques~\cite{zhang2025towards,zhou2025dense360}.
However, the existing works prefer to directly fine-tune the multi-modal large language models based on the existing VQA datasets. 
Thus, works like ERP-RoPE~\cite{zhou2025dense360} attempt to explore the internal features of panoramic images, which further enhance the model's understanding of panoramic images.

\subsection{Dataset}
Moreover, datasets also serve as the solid foundation of the three above tasks. We categorize omnidirectional datasets into three domains: Indoor, Outdoor, and UAV/Flight, each further divided into real or synthetic sources.
From this taxonomy, which is shown in Table~\ref{tab:omni-dataset}, we identify 23 representative datasets, providing a structured basis for comparing scale, modalities, and annotations.
    
Across these directions, common modalities include RGB panoramas, depth, camera poses, and semantic labels. 
Datasets differ in granularity: indoor resources often offer dense semantics, scan-based or synthetic sets yield geometry and object metadata~\cite{chang2017matterport3d}, while outdoor or aerial captures prioritize spatial coverage over instance detail~\cite{huang2023360vot}. 
Temporal scope, viewpoint density, and annotation type further diversify the landscape.

Recent datasets increasingly combine geometric ground truth with downstream tasks. For example, Matterport3D~\cite{chang2017matterport3d}, ReplicaPano~\cite{dong2024panocontext}, and HM3D~\cite{ramakrishnan2021habitat} illustrate datasets that pair panoramas/depth/layouts with instance or task-level labels. Temporal and embodied captures are becoming more common~\cite{Xia_2018_CVPR}. However, fully synchronized multi-sensor panoptic collections that combine panoramic RGB, LiDAR, spatial audio, and behavioral traces remain rare~\cite{kim2024pair360}. Researchers are paying more attention to projection-induced distortions and adopting distortion-aware methods or losses to mitigate latitude or pole artifacts~\cite{zhang2022bending}. Finally, QA-style panoramic reasoning benchmarks have begun to appear (e.g., OSR-Bench~\cite{dongfang2025multimodal} and OmniVQA~\cite{zhang2025towards}), indicating early interest in reasoning-oriented panoramic evaluation even though these resources are still nascent.
    
\section{PANORAMA System Architecture} 
\begin{figure}
        \centering
        \includegraphics[width=1\linewidth]{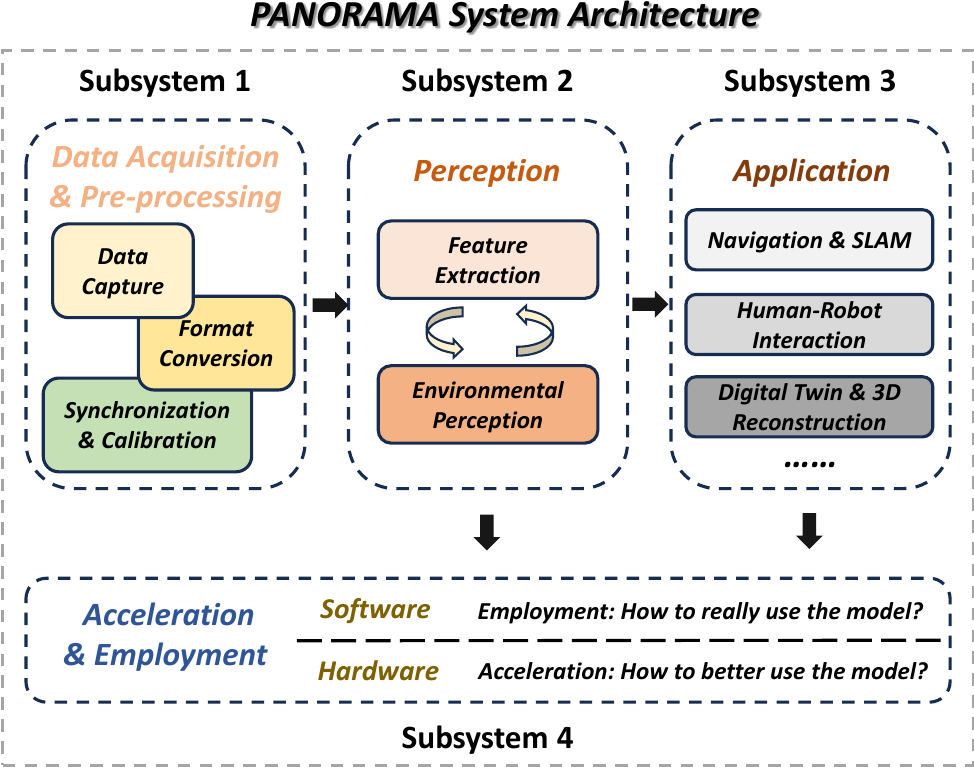}
        \caption{The overview of the system architecture.}
        \label{fig:figure2}
    \end{figure}

The advent of embodied intelligence presents fundamental challenges and opportunities to omnidirectional vision.
Therefore, a dedicated panoramic system is not merely beneficial but necessary to unlock the potential of omnidirectional vision for embodied AI.
Thus, as shown in Figure~\ref{fig:figure2}, we propose an ideal panoramic system, \textit{\textbf{PANORAMA}}, comprised of four integrated subsystems, which has the potential to be a key solution for the integration of panoramic vision and embodied AI.
\subsection{Key Subsystems}
\subsubsection{Subsystem 1: Data Acquisition \& Pre-processing}
This subsystem is responsible for capturing raw omnidirectional data and converting it into a format suitable for computational processing. 
It primarily consists of hardware like cameras (e.g., those utilizing equirectangular projection or multi-fisheye lens rigs) and complementary sensors (e.g., IMUs, depth sensors). 
Its core functions include:
\begin{itemize}
    \item \textbf{Data Capture:} Acquiring high-resolution omnidirectional images and videos.
    \item \textbf{Format Conversion:} Dynamically transforming data between different representations (e.g., ERP, Cubemap) to suit the needs of downstream processing tasks.
    \item \textbf{Synchronization \& Calibration:} Ensuring temporal alignment and spatial coordination between multiple/multi-modal sensors for accurate data fusion.
    \end{itemize}
    
    \subsubsection{Subsystem 2: Perception}
    This subsystem performs fundamental scene perception on the preprocessed panoramic data.
    It employs deep learning models adapted for spherical geometry to extract rich, structured information from the omnidirectional input.
    Its key capabilities include:
    \begin{itemize}
        \item \textbf{Feature Extraction:} Utilizing specialized architectures (e.g., Spherical CNNs, Transformers) to understand the omnidirectional context.
        \item \textbf{Environmental Perception:} Simultaneously performing core perception tasks such as semantic segmentation, object detection, and depth estimation from a shared feature backbone for efficiency.
    \end{itemize}

    \subsubsection{Subsystem 3: Application}
    This subsystem translates the perceptual insights into actions for embodied AI agents. 
    It consumes the structured data (e.g., semantic maps, object lists, depth information) to serve specific downstream tasks. 
    Example applications include:
    \begin{itemize}
        \item \textbf{Navigation \& SLAM:} Enabling autonomous movement and real-time spatial mapping in indoor and outdoor environments.
        \item \textbf{Human-Robot Interaction:} Providing agents with full-scene awareness for more natural and context-aware interactions.
        \item \textbf{Digital Twin \& 3D Reconstruction:} Creating immersive and accurate virtual models of real-world spaces for simulation, planning, and monitoring.
    \end{itemize}
    
    \subsubsection{Subsystem 4: Acceleration \& Employment}
    This subsystem addresses the computational challenges of processing high-resolution panoramic data in real-world, often resource-constrained settings. 
    It focuses on the practical implementation of the entire pipeline.
    \begin{itemize}
        \item \textbf{Software Acceleration:} Optimizing the entire stack through techniques like model quantization and pruning to balance accuracy, latency, and power consumption for deployment on edge devices.
        \item \textbf{Hardware Employment:} Employing edge computing platforms (e.g., NVIDIA Jetson, SOPHGO SE9) to achieve real-world applications.
    \end{itemize}


    \subsection{Workflow}
    The whole system operates as a cohesive and integrated pipeline. 
    The workflow begins with the \textbf{Data Acquisition \& Pre-processing Subsystem}, where raw data from panoramic cameras and other sensors is captured, corrected, and synchronized. 
    This clean, formatted data is then passed to the \textbf{Perception Subsystem}, where deep learning models perform feature extraction and environmental perception to generate a comprehensive understanding of the scene. 
    These perceptual outputs are subsequently utilized by the \textbf{Application Subsystem} to execute specific embodied AI tasks, such as navigation or interaction. 
    Throughout this process, the \textbf{Acceleration \& Deployment Subsystem} ensures the computational feasibility of the pipeline, enabling low-latency, efficient operation on edge devices from raw sensor inputs to final embodied applications.

    \section{Emerging Trends \& Future Roadmap}

    \begin{figure*}[th!]
        \centering
        \includegraphics[width=0.9\textwidth]{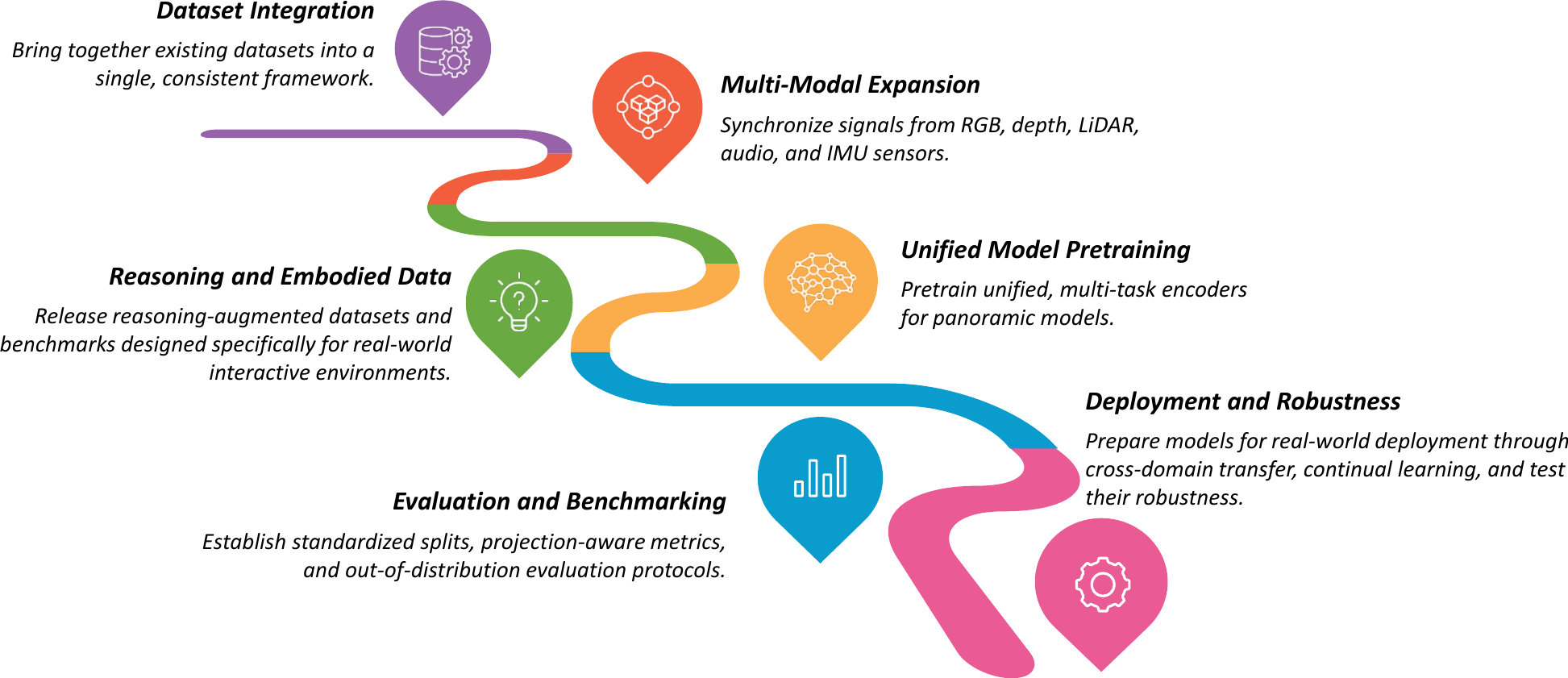}
        \caption{
        Roadmap for implementing an omnidirectional model in the Embodied AI area. 
        }
        \label{fig:roadmap}
    \end{figure*}
    
    
Recent advancements in omnidirectional vision have shown promising progress across various downstream applications, including visual perception~\cite{zheng2023both}, reasoning~\cite{dongfang2025multimodal}, and embodied tasks such as navigation~\cite{wang2025panogen++} and grasping~\cite{kerr2025eye}. Representative model architectures often involve panoramic extensions of 2D CNNs or Transformer backbones, as well as unified multi-task encoders trained on 360-degree geometry and semantics~\cite{sun2021hohonet, jiang2021unifuse}. Additionally, early multi-modal models have emerged that integrate vision and language for more comprehensive task understanding. Despite these advancements, many approaches remain task-specific, struggle with projection ambiguities, and lack large-scale multi-modal pretraining resources. These limitations hinder model generalization and pose significant challenges to the broader development of embodied AI~\cite{he2022rethinking, zhang2025towards}.
    To overcome this gap and realize the PANORAMA system, we further propose a staged roadmap as shown in Figure~\ref{fig:roadmap} to build an ideal unified model for omnidirectional tasks. The timeline spans six stages:
    
    \subsubsection{Stage 1: Dataset Integration}

    In the first stage, the focus is on bringing together existing datasets into a single, consistent framework for projections, along with standardized test splits. The data will be re-annotated with consistent labels, and flexible re-projection tools will help ensure fair comparisons of model performance across different formats, like ERP and cubemap. This stage will result in a well-organized benchmark suite, with careful human checks to reduce errors in the annotations and improve label accuracy.

    \subsubsection{Stage 2: Multi-Modal Expansion}

    Next, the focus shifts to synchronizing signals from RGB, depth, LiDAR, audio, and IMU sensors, enabling multi-modal and multi-task pretraining tailored for panoramic vision. Standardized rigs and calibration protocols will facilitate richer sensor fusion, enhancing the modeling of environments captured by panoramic cameras. A key milestone in this phase would be the creation of a public multi-sensor corpus with harmonized splits, enabling more effective benchmarking. To offset the cost of large-scale data collection, hybrid real–synthetic pipelines will be utilized, combining both real-world and simulated data for more robust sensor training.
    

    \subsubsection{Stage 3: Reasoning and Embodied Data}

The third stage focuses on advancing reasoning capabilities in interactive, embodied tasks such as grounded visual question answering (VQA), instruction-following, navigation, and grasping. These tasks require robust spatial reasoning to understand and interact with the environment. To facilitate this, hybrid question generation methods—combining templates, large language models (LLMs), and human validation—will be employed to ensure both scale and diversity in the training data. Simulation environments will play a crucial role in providing varied and dynamic scenarios for tasks like navigation and grasping, where precise spatial awareness and decision-making are essential. The release of a reasoning-augmented dataset and a benchmark designed specifically for evaluating spatial reasoning, navigation, and grasping performance will establish standardized protocols for measuring the success rates of models in real-world interactive environments.


\subsubsection{Stage 4: Unified Model Pretraining}

Building on the integrated multi-modal corpora from earlier stages, this phase focuses on pretraining unified, multi-task encoders for panoramic models. These models jointly process 360$^{\circ}$ geometry, semantic labels, and synchronized sensor streams (RGB, depth, LiDAR, audio, and IMU). 
Based on weights from established 2D or 3D architectures, the key innovation is fine-tuning and post-training using panoramic-specific datasets and task objectives. Training incorporates cross-projection representations, multi-objective losses, and domain-mixing curricula (real–synthetic) to ensure transferability. This stage adapts traditional models to the challenges of panoramic vision and enhances generalization in real-world scenarios like navigation, grasping, and embodied tasks. Fine-tuning on tailored datasets ensures efficient learning for both structured and unstructured data in dynamic environments.

    \subsubsection{Stage 5: Evaluation and Benchmarking}

    This stage establishes a rigorous evaluation infrastructure consisting of standardized dataset splits, projection-consistent reprojection tools, and a harmonized metric suite covering per-task accuracy, cross-projection consistency, and success rates of reasoning and embodiment tasks. Protocols include explicit out-of-distribution splits, calibration and uncertainty measures, efficiency targets, and human-verified evaluation for critical tasks; together, these components enable reproducible comparison, ablation studies, and operational readiness assessments.


    \subsubsection{Stage 6: Deployment and Generalization}

In the final stage, the focus is on preparing models for real-world deployment through cross-domain transfer, continual learning, and testing their robustness. Models are tested under real-world conditions, including changes in data distributions, using out-of-distribution (OOD) splits. Evaluation will include measures like calibration, latency, and uncertainty. This stage also includes delivering a deployment kit with stress-test datasets, evaluation benchmarks for ongoing adaptation, and workflows to validate models’ uncertainty.

    \section{Cross-Community Impacts \& Open Challenges} 
    In today's era of embodied AI, the maturity of omnidirectional vision no longer merely refers to the gradual update of technology but rather its emergence as a fundamental enabling technology, promoting cross-community breakthroughs in practical application domains. 
    The impact of omnidirectional vision, especially the PANORAMA system, in the current era extends far beyond a single community, providing cross-community impacts for researchers from diverse fields:
    \begin{itemize}
        \item \textbf{Robotics \& Autonomous Navigation:} For mobile robots and autonomous vehicles, omnidirectional perception is the cornerstone of complete situational awareness.
        It eliminates blind spots, making navigation in dense and dynamic environments—such as crowded public places—more accurate and safer~\cite{ran2017convolutional,huang2024360loc}, enhancing the robot's perception ability by providing contextual information from different angles~\cite{park2024360}.

        \item \textbf{Human-Robot Interaction:} Omnidirectional vision enables robots to understand social and spatial information like humans. A robot equipped with an omnidirectional camera can simultaneously track multiple individuals~\cite{huang2024360loc,li2024omnissrzeroshotomnidirectionalimage}, interpret group conversations, and comprehend social cues from any direction~\cite{rai2017omini,martin2022scangan360}, thereby facilitating more natural, seamless, and trustworthy human-robot interaction.

        \item \textbf{Cognitive AI \& Virtual Agents:}  Omnidirectional vision provides a dense, information-rich perceptual stream that is fundamentally closer to a human's egocentric vision of the world~\cite{Xia_2018_CVPR,zhou2025dense360}. This high-fidelity input is crucial for developing the foundation of high-level, human-like cognitive abilities, including spatial reasoning, long-horizon task planning, and commonsense understanding of environmental physics.

    \end{itemize}

    Despite the positive cross-community impacts of omnidirectional vision in the embodied AI era, several open challenges remain, providing new directions for future research.

    \begin{itemize}
        \item \textbf{Generalization and Robustness:} Most current models still focus on specific scenarios or projection methods~\cite{ai2022deeplearning}. Developing models that generalize across diverse panoramic sensor specifications, application scenarios, and projection methods remains non-trivial~\cite{ai2025survey}. It is necessary for future works to focus on projection-agnostic representations and self-supervised learning techniques that can learn invariant features from unlabeled omnidirectional information, including both images and video streams.

        \item \textbf{Dynamic Distortion Handling:} While current approaches have made significant progress in handling the static distortion of panoramic images~\cite{zhang2022bending}, they treat it as a frame-independent geometric problem. This represents a critical limitation, as the distortion is fundamentally dynamic in real-world scenarios. Future research should further explicitly consider the temporal consistency and evolution of distortions across omnidirectional video sequences.

        \item \textbf{Action-Aware Representation Learning:} The ultimate goal of omnidirectional vision in the era of embodied AI is not merely to enable robots to observe better, but to empower them to take action more effectively~\cite{xu2024survey,xu2025embodied}. A key direction is to allow models to learn action-oriented representations within panoramic images~\cite{mu2023embodiedgpt,fu2025segman}. By integrating the unique advantages of omnidirectional visual features into downstream control strategies, we can achieve more effective and efficient decision-making in robotic behavior.

        \item \textbf{Scalable and Unified Architectures:} An important challenge is the creation of unified, multi-task foundation models specifically designed for omnidirectional vision~\cite{dong2024panocontext,shen2022panoformer}. Moving beyond the inefficiency of task-specific models, these models would be pre-trained on extensive panoramic data to capture a fundamental understanding of omnidirectional geometry and semantics~\cite{sun2021hohonet,jiang2021unifuse}. This would yield a powerful visual backbone that could be rapidly specialized for numerous applications, improving performance and generalization while reducing the need for massive task-specific datasets.

    \end{itemize}

    In summary, the process towards achieving truly embodied intelligence is a collective effort. In the field of omnidirectional vision, we call upon researchers to:

    \begin{itemize}
        \item \textbf{For Dataset Creators:} Plan and publish large-scale multi-task omnidirectional datasets that encompass the complexity of real-world scenes, including both indoor and outdoor scenarios, general scenarios, and embodied intelligent scenarios.

        \item \textbf{For Algorithm Researchers:} Go beyond simple adaptations based on pinhole models and create novel architectures and dynamic learning paradigms that possess omnidirectional information,  which is the key to embracing the unique challenges of omnidirectional vision.

        \item \textbf{For Application Engineers:} Explore and demonstrate the benefits of omnidirectional perception in real-world robotics and interactive systems, as it bridges the gap between laboratory research and practical applications.

    \end{itemize}

The era of embodied intelligence demands a paradigm shift from a narrow, forward-looking view to a comprehensive, spherical understanding of our world. By embracing omnidirectional vision, we can collectively unlock the next frontier of embodied AI, creating agents that not only see but also understand and interact with the entirety of their environment. \textit{The embodied AI's future is omnidirectional.}

\appendix

\bibliography{aaai2026}

\end{document}